\documentclass{article}

 \usepackage[preprint]{neurips_2026}

\usepackage[utf8]{inputenc} 
\usepackage[T1]{fontenc}    
\usepackage{hyperref}       
\usepackage{url}            
\usepackage{booktabs}       
\usepackage{amsfonts}       
\usepackage{nicefrac}       
\usepackage{microtype}      
\usepackage{xcolor}         

\definecolor{pastelblue}{RGB}{173, 216, 230}
\definecolor{pastelgreen}{RGB}{152, 251, 152}
\definecolor{pastelorange}{RGB}{255, 200, 150}
\definecolor{pastelred}{RGB}{255, 182, 193}
\definecolor{pastelpurple}{RGB}{221, 160, 221}
\definecolor{pastelyellow}{RGB}{253, 253, 150}
\definecolor{pastelpink}{RGB}{255, 192, 203}

\usepackage{tikz}
\usetikzlibrary{positioning, fit, backgrounds, arrows.meta, shapes.geometric, decorations.pathreplacing}

\usetikzlibrary{shapes.geometric, shapes.multipart, arrows.meta, positioning, fit, backgrounds, decorations.pathreplacing, decorations.pathmorphing, shadows, calc}

\definecolor{pastelblue}{RGB}{198, 219, 239}
\definecolor{pastelblue2}{RGB}{148, 219, 254}
\definecolor{pastelgreen}{RGB}{199, 233, 192}
\definecolor{pastelorange}{RGB}{199, 233, 192}
\definecolor{pastelred}{RGB}{252, 146, 114}
\definecolor{ndarkgray}{RGB}{100, 100, 100}
\definecolor{nblue}{RGB}{31,119,180}
\definecolor{nred}{RGB}{214,39,40}
\definecolor{ngreen}{RGB}{44,160,44}
\definecolor{norange}{RGB}{255,127,14}

\definecolor{nblue}{RGB}{31,119,180}
\definecolor{norange}{RGB}{255,127,14}
\definecolor{ngreen}{RGB}{44,160,44}
\definecolor{nred}{RGB}{214,39,40}
\definecolor{npurple}{RGB}{148,103,189}
\definecolor{ngray}{RGB}{220,220,220}
\definecolor{ndarkgray}{RGB}{120,120,120}
\definecolor{pastelyellow}{RGB}{255,235,153}

\usepackage{tabularx}
\usepackage{multirow}%
\newcolumntype{L}[1]{>{\hsize=#1\hsize\raggedright\arraybackslash}X}
\newcolumntype{C}[1]{>{\hsize=#1\hsize\centering\arraybackslash}X}
\newcolumntype{R}[1]{>{\hsize=#1\hsize\raggedleft\arraybackslash}X}

\usepackage{amsmath,amssymb,amsfonts}%
\usepackage{mathrsfs}%
\usepackage{algorithmicx}%
\usepackage{algpseudocode}%
\usepackage{listings}%
\usepackage{todonotes}%
\usepackage[depth=2]{bookmark}
\usepackage{caption}
\usepackage{subcaption}
\usepackage{float}

\newcommand{\figpanel}[2]{Figure~\ref{#1}#2}

\usepackage{color-edits}
\addauthor{gemini}{red}
\addauthor{jakub}{blue}
\addauthor{gustav}{magenta}

\title{Incremental Evaluation and Training\\in Relational Deep Learning}

\author{
  Jakub Peleška \\
  Czech Technical University in Prague\\
  \texttt{jakub.peleska@cvut.cz}
  \And
  Gustav Šír \\
  Czech Technical University in Prague\\
  \texttt{gustav.sir@cvut.cz} 
}

\begin{document}

\maketitle

\begin{abstract}

Relational Deep Learning (RDL) models multi-tabular databases as temporal heterogeneous graphs to enable end-to-end representation learning. However, prevailing RDL evaluation practices rely on static, single-episode dataset snapshots, overlooking the continuous, time-evolving nature of real-world databases. 
Consequently, current RDL benchmarks fail to capture how model performance changes as new data accumulates over time.
To address this limitation, we introduce an incremental, multi-episode evaluation and training paradigm to assess and improve the temporal robustness and adaptability of state-of-the-art RDL models.
Using established large-scale datasets, we examine data evolution and model training dynamics, demonstrating that temporal concept drifts occur in the majority of predictive tasks. We present multiple incremental training regimes for fine-tuning the models and demonstrate that transfer learning is both feasible and highly effective in the RDL setting.
Alongside a new temporal evaluation metric that prioritizes near-future accuracy, we show that our incrementally fine-tuned models consistently outperform the standard, expensive, from-scratch trained baselines.

\end{abstract}



\section{Introduction}
\label{sec:intro}

\begin{figure}[t]
    \centering
    \resizebox{\textwidth}{!}{
        \input{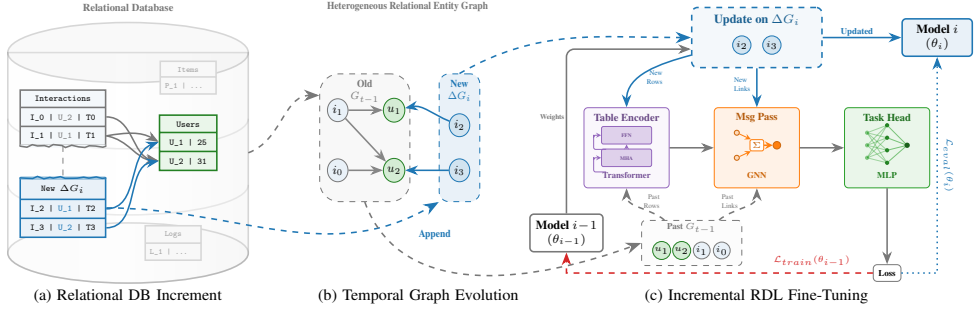}
    }
    \caption{Incremental RDL workflow with (a) append-only data updates, (b) induced relational graph increments $\Delta G_i$ on top of historical context $G_{t-1}$, and (c) episode-wise fine-tuning from $\theta_{i-1}$ to $\theta_i$.}
    \label{fig:continuous_RDL}
\end{figure}

\textit{Relational databases (RDBs)} are the foundational infrastructure for data storage across a vast majority of modern industries~\cite{Codd1970, agrawal2001storage, white_pubmed_2020}. Despite their ubiquity and immense predictive value, applying traditional machine learning (ML) to RDBs presents significant challenges. The inherently heterogeneous, dynamically structured nature of relational data—spanning multiple interconnected tables with varying schemas—resists the fixed-size feature vector representations required by classical ML algorithms. Consequently, practitioners are often forced into laborious manual feature engineering or lossy table joins that discard important structural semantics~\cite{fey2024position}.
To address these bottlenecks, \textit{Relational Deep Learning (RDL)}~\cite{zahradnik2023deep, fey2024position} has recently emerged as a transformative ML paradigm. By interpreting an entire relational database as a heterogeneous graph, RDL leverages Graph Neural Networks (GNNs)~\cite{velivckovic2018graph, brody2022how} alongside deep tabular representation learning~\cite{gorishniy_revisiting_2021, peleska_transformers_2025} to enable end-to-end predictive modeling directly on the native structures of the RDBs~(\figpanel{fig:continuous_RDL}{}).

While RDL successfully eliminates the need for manual feature flattening, bringing ML significantly closer to the raw, heterogeneous reality of relational data, it currently neglects a foundational dimension of real-world RDBs: their continuous temporal evolution. Indeed, databases are not static repositories; they dynamically expand and shift over time, causing models trained on historical snapshots to become incrementally obsolete. Despite this dynamic reality, current RDL evaluation practices remain focused on static models. Given the structural complexity of RDBs, which prevents classic i.i.d. sample-based predictions, predictive tasks in RDL are defined by augmenting the schema with a dedicated ``task table'' that specifies target entities and labels. To avoid temporal leakage, existing frameworks~\cite{robinson2024relbench, peleska_redelex_2025} correctly use timestamps to perform chronological train–test splits~\cite{fey2024position,ranjan_relational_2025}. However, this snapshot-based evaluation assesses models in a single, isolated episode, fundamentally failing to capture how model performance degrades over time, e.g., due to temporal concept drift. Characterizing this temporal stability and evaluating how models adapt to new increments of relational data then remains an unaddressed prerequisite for advancing toward robust, real-world RDL foundation models~\cite{ranjan_relational_2025, wang_griffin_2025}.


\paragraph{Contributions.} To address the gap between static benchmarking and the dynamic reality of relational data, we introduce an incremental, multi-episodic evaluation and training paradigm, visualized in Figure~\ref{fig:continuous_RDL}. Tailored for robust assessment of RDL models, our approach simulates the real-world life-cycle of database growth by transitioning RDL benchmarking from static snapshots to continuous, horizon-shifting evaluation. Leveraging state-of-the-art models and large-scale datasets~\cite{robinson2024relbench, peleska_redelex_2025, gu_relbench_2026}, we empirically demonstrate that temporal concept drift is a prevalent factor in standard RDL tasks. To combat model degradation, we define and benchmark multiple incremental fine-tuning strategies, showing that knowledge transfer is highly effective and computationally superior to from-scratch retraining in the continuous RDL setting. Finally, we propose a suitable exponential decay-based evaluation that prioritizes near-future predictive accuracy, closely aligning benchmark evaluation with practical operational priorities.

\section{Related Work}
\label{sec:related_work}

\paragraph{RDL Benchmarks and Frameworks}
Standardized ecosystems have accelerated RDL research. RelBench~\cite{robinson2024relbench, gu_relbench_2026} formalized predictive tasks with chronological splits to prevent temporal leakage~\cite{kocijan_predictive_2026}. Concurrently, ReDeLEx~\cite{peleska_redelex_2025} expanded the evaluation landscape across a large set of diverse databases to analyze structural impacts on model performance. While these frameworks provide essential building blocks for static testing, they lack the multi-episodic evaluation necessary to measure continuous model degradation and real-world lifecycle dynamics.

\paragraph{Architectural Advances and Foundation Models}
To capture the extreme heterogeneity of RDBs, many recent methods integrate tabular representation learning~\cite{lachi_boosting_2025, peleska_transformers_2025}, LLMs~\cite{wu_large_2025}, and graph transformers~\cite{dwivedi_relational_2025, lachi_rgp_2025}. This progress naturally steers the field toward RDL foundation models~\cite{wehrstein_towards_2025, dwivedi_challenges_2025, fey_kumorfm_2025, hudovernik_kumorfm-2_2026, wang_griffin_2025, ranjan_relational_2025}, supported by task-agnostic pretraining~\cite{peleska_task-agnostic_2025} and synthetic data generation~\cite{jurkovic2025syntherela, kothapalli_plurel_2026}. While these foundational architectures are largely designed for static deployment, the overarching shift towards foundation models and transfer learning strongly overlaps with our proposed temporal fine-tuning, forming a prerequisite for maintaining their robustness over time.

\paragraph{Temporal Graph Representation Learning}
Relational data evolution conceptually intersects with Temporal Graph Representation Learning~\cite{kazemi2020representation}; nevertheless, traditional temporal GNNs (e.g., TGN~\cite{rossi2020temporal}, TGAT~\cite{da2020time}) model high-frequency dynamic systems to predict immediate interactions. In contrast, our incremental RDL paradigm does not treat databases as stateful time-series systems. Instead, we focus on macro-level volumetric growth and long-term temporal concept drift to assess model obsolescence over extended horizons.

\paragraph{Continual Learning and Concept Drift}
By addressing model obsolescence, our approach conceptually bridges RDL with Continual Learning (CL)~\cite{parisi2019continual} and concept drift adaptation~\cite{lu2018learning}. Standard CL and Continual Graph Learning~\cite{zhang2022continual} mitigate catastrophic forgetting primarily in homogeneous structures. RDL, however, involves highly heterogeneous, multi-tabular data increments. Our work provides the first formal benchmark to evaluate CL strategies directly within the dynamically evolving environment of real-world databases.




\section{Relational Deep Learning}
\label{sec:rdl}

RDL is a machine learning paradigm built on interpreting relational databases as heterogeneous graphs. This perspective enables the direct application of deep learning techniques---specifically GNNs alongside tabular representation learning---on the native structures of RDBs. While the underlying idea was explored in earlier work~\cite{cvitkovic2020supervised, zahradnik2023deep}, it was recently formalized under the RDL umbrella~\cite{fey2024position} to consolidate these efforts into a cohesive research field, detailed below. 

\subsection{Graph Representation} 
Formally, a {relational database} $\mathcal{R}$ is defined as a finite set of relations $R_1, R_2, \dots ,R_n$. Each relation $R$ consists of a heading (signature) $R_{/n}$, formed by the set of its attributes, and a body, formed by the tuples of attribute values ${t_i} = (a_1, a_2, \ldots, a_n)$, commonly represented as a \textit{table} $T_R$. Additionally, the relation $R$ can be defined extensionally by the \textit{unordered} set of its tuples: $R = \{t_1, t_2,\ldots, t_m\}$, corresponding to the \textit{rows} (or entities) of the table $T_R$.

Practically, the attributes forming the rows of each table $T_R$ are categorized into four functional groups: (i) \textit{primary key} (PK), a minimal set of attributes $PK_R$ that uniquely identifies each entity, such that for any two distinct tuples $t_a, t_b \in R$, $t_a[PK_R] \neq t_b[PK_R]$; (ii) \textit{foreign key} (FK), a set of attributes in a relation $R_i$ that references the primary key of another relation $R_j$, establishing a structural dependency ensuring referential integrity $\forall t \in R_i, t[FK_{R_i \to R_j}] \in \{t'[PK_{R_j}] \mid t' \in R_j\}$; (iii) \textit{feature attributes}, containing characteristic information about the entity; and, optionally, (iv) a \textit{timestamp}, indicating when the row was created or last modified.

The \textit{relational entity graph} of a database $\mathcal{R}$ is a \textit{temporal heterogeneous graph} defined as $G = (\mathcal{V}, \mathcal{E}, \phi, \psi, \tau)$. The set of nodes $\mathcal{V}$ is defined as the union of all tuples from each relation: $\mathcal{V} = \{v_{i,j} \mid R_i \in \mathcal{R}, t_j \in R_i\}$, inferring $v_{i,j} \simeq t_j \in R_i$. Consequently, the set of edges $\mathcal{E}$ is defined by the PK-FK relationships: $\mathcal{E} = \{(v_{i,k}, v_{j,l}) \mid t_k \in R_i, t_l \in R_j, t_k[FK_{R_j}] = t_l[PK_{R_j}] \lor t_l[FK_{R_i}] = t_k[PK_{R_i}]\}$~(\figpanel{fig:continuous_RDL}{b}).

To capture the heterogeneity of the database, the function $\phi: \mathcal{V} \to \mathcal{T}^v$ serves as a \textit{node mapping}. The set of node types $\mathcal{T}^v$ corresponds directly to the set of relations within $\mathcal{R}$, meaning $\phi$ maps a node $v_{i,j}$ strictly to its original source relation $R_i$. Similarly, $\psi: \mathcal{E} \to \mathcal{T}^e$ is an \textit{edge mapping} function, where the set of edge types $\mathcal{T}^e$ corresponds to the specific pairs of PK and FK attributes (or their reversed counterparts) that generated the edge. Finally, the \textit{time mapping} function $\tau: \mathcal{V} \to \mathrm{T}$ assigns a timestamp to each node such that $\tau(v) = \mathrm{T}_v$. If a temporal attribute is not available for a given row, a zero timestamp is assigned by default.


\subsection{Neural Methods}

Building upon the graph representation $G$, RDL models typically follow a four-stage pipeline: (i) a \textit{table-level attribute encoder} initializes node embedding matrices $h_v^{(0)} \in \mathbb{R}^{d^{(0)} \times n}$ by encoding each feature attribute $A_1,\dots,A_n$ of the node's source relation $\phi(v)$ based on its semantic data type; (ii) a \textit{table-level tabular model} employs tabular architectures~\cite{chen2023trompt, hu2020tabtransformer} to refine these embeddings into $h_v^{(l)} \in \mathbb{R}^{d^{(l)} \times n}$, optionally compressing the matrix into a single vector $h_v^{(l)} \in \mathbb{R}^{d^{(l)}_{\phi(v)}}$; (iii) a \textit{graph neural model} applies message-passing based on the defined edge types $\mathcal{T}^e$, using standard GNNs~\cite{Hamilton2017InductiveGraphs, velivckovic2018graph, brody2022how} for vectors or custom schemes~\cite{peleska_transformers_2025, chen_relgnn_2025} for attribute-matrices; and (iv) a \textit{task-specific model head} maps the final node embeddings to predictions, typically via simple MLP layers. This RDL architectural blueprint is illustrated in \figpanel{fig:continuous_RDL}{c}.

\subsection{Predictive Tasks}
\label{sec:rdl-tasks}
Predictive tasks in RDL are formalized by constructing a dedicated ``{task table}'' $T_{task}$ that extends the existing relational database $\mathcal{R}$~\cite{fey2024position, kocijan_predictive_2026}.
A task table $T_{task}$ consists of tuples defining the individual prediction instances, abstracted as $(v, y, t)$. Here, $v \in \mathcal{V}$ is the target entity (node) identified via a foreign key reference $T_{task}[FK]$, $y \in \mathcal{Y}$ is the ground-truth target label to be predicted (e.g., $y \in \mathbb{R}$ for regression), and $t \in \mathrm{T}$ is the \textit{anchor time}, specifying the moment at which the prediction is made.

The objective of an RDL model $f_\theta$ is to predict the target label $\hat{y} = f_\theta(G_{v, \leq t}) \approx y$, where $G_{v, \leq t}$ is the relational subgraph defining the computational context for entity $v$, strictly filtered to only include information available at or before anchor time $t$. Depending on the role of $t$, these tasks are categorized into static and temporal formulations.

\paragraph{Static Tasks.}
Static tasks involve predicting missing values or properties within a database viewed as a single, fixed snapshot in time. The temporal dimension is effectively ignored, reducing instances to pairs $(v, y)$. The objective simplifies to learning a function $\hat{y} = f_\theta(G_v)$, utilizing the entire available graph $G$ without risk of temporal information leakage. In these cases, the target label $y$ belongs to a feature column that already structurally exists within the database but is masked out.

\begin{figure}[t]
    \centering
    \resizebox{\textwidth}{!}{%
        \input{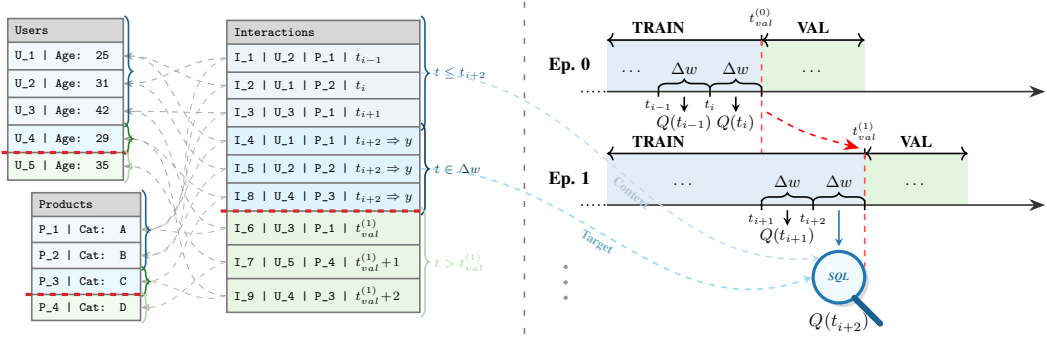}
    }
    \caption{
    Left: a relational task table is chronologically split by anchor times, with a leakage barrier (red) separating validation from training context, further split into respective target task windows of size $\Delta w$, induced by the predictive queries $Q(t_i)$. Right: evaluation proceeds over increments of length $\Delta I$, shifting the train/validation horizon $t^{(i)}_{val}$ and instantiating the queries at successive anchors, while extracting context $G_{\le t}$ and future targets from the newly revealed task windows.
    }
    \label{fig:temporal-tasks}
\end{figure}

\paragraph{Temporal Tasks.}
Conversely, \textit{temporal tasks} (or ``{forecasting tasks}'') predict future behaviors or events that have not yet occurred. Here, the anchor time $t$ is crucial, as the label $y$ represents an aggregation or occurrence in a future window $(t, t + \Delta w]$ (e.g., total sales over the next 30 days). The set of task labels is formed as $Y = \{Q_{\Delta w}(t) \mid t \in \mathrm{T}_{task}\}$, where $Q$ is a future-pointing task query~\footnote{The queries are implemented through a SQL~\cite{chamberlin_sequel_1974} representation of the task.} with a fixed window size $\Delta w$, and $\mathrm{T}_{task}$ is a set of task anchor timestamps (see Figure~\ref{fig:temporal-tasks}). To prevent temporal data leakage, the model must strictly condition its representations on past data, requiring the graph to be temporally masked to yield $G_{\leq t} = (\mathcal{V}_{\leq t}, \mathcal{E}_{\leq t})$, where $\mathcal{V}_{\leq t} = \{u \in \mathcal{V} \mid \tau(u) \le t\}$.

\paragraph{Data Splits.}
The strategy for partitioning data varies fundamentally between the tasks. For static tasks, test instances $(v, y)$ can be split randomly. Temporal tasks, however, necessitate strict \textit{chronological evaluation} to prevent backward data leakage (Figure~\ref{fig:temporal-tasks}). These tasks further define preset validation and test timestamps: the training set contains instances where $t < t_{val}$, the validation set where $t_{val} \le t < t_{test}$, and the test set where $t \ge t_{test}$. This ensures the model is evaluated exclusively on unseen future time periods.

\section{Incremental Evaluation Paradigm}
\label{sec:incr-eval}
Standard evaluation of models on relational databases relies on static dataset snapshots, failing to capture the dynamic evolution of real-world data where temporal increments dictate both target creation and predictive tasks. To standardize the measurement of model robustness against undetected data shifts and temporal concept drifts, we propose an incremental evaluation paradigm that simulates the real-world deployment lifecycle (Figure~\ref{fig:continuous_RDL}). This enables researchers to assess how well RDL models adapt to continuously growing databases over time.

\subsection{Multi-Episodic Relational Tasks}
\label{sec:mult-ep-rdl-tasks}
Expanding upon the single-episode temporal tasks defined by future-pointing queries $Q_{\Delta w}(t)$ (Section~\ref{sec:rdl-tasks}), we introduce a fixed time increment $\Delta I \ge \Delta w$ representing the temporal stride between consecutive model updates (Figure~\ref{fig:temporal-tasks}), where an \textit{episode} constitutes a batch of new target data arriving over this increment.

To transition to \textit{multi-episodic} setting, we evaluate and update models incrementally over a sequence of monotonically increasing evaluation horizons: $T_{\Delta I} = \{t_i \mid t_i = t_{i-1} + \Delta I, \quad t_i \le t_{val}, \quad i \ge 1\}$.

At each episode $i$, new training examples generated by $Q_{\Delta w}(t)$ from newly available anchor times $t \in (t_{i-1}, t_i]$ are revealed up to the anchor time $t_i$. This interval may contain multiple base task windows $\Delta w$. Once evaluated, the true labels become available, and the increment is appended to the historical training data. This strict \textit{append-only growth} (\figpanel{fig:continuous_RDL}{a}) prevents temporal leakage, ensuring the model never accesses information beyond its current validation horizon.

\subsection{Incremental Training Regimes}
\label{sec:incr-train-regime}


As the prediction horizon shifts from episode $i-1$ to episode $i$ (Figure~\ref{fig:temporal-tasks}), a new data increment becomes available. The training set correspondingly expands to include all data up to the new validation timestamp $t^{(i)}_{val}$. The validation and test datasets shift forward chronologically, maintaining their structural role but evaluating the model on the newly established horizon. Because an increment can span multiple task windows, these shifted splits dynamically adapt to test the model on the exact block of newly revealed target data.

To evaluate how effectively models assimilate these increments, we benchmark four training regimes:
\begin{enumerate}
    \item \textit{From Scratch (Baseline):} Trained \textit{from scratch} using all available data up to $t^{(i)}_{val}$, serving as a (computationally expensive) baseline that follows standard single-episode RDL training.
    \item \textit{Finetune (Cumulative):} Retains weights from episode $i-1$ and \textit{fine-tunes} on the entire historical dataset up to $t^{(i)}_{val}$.
    \item \textit{Finetune (Incremental):} Updates the episode $i-1$ model \textit{exclusively} on the newest data increment to maximize training speed and immediate responsiveness.
    \item \textit{Finetune (Upsampled):} A hybrid strategy fine-tuning on a distribution that \textit{upsamples} the most recent increment while retaining historical data to mitigate \textit{catastrophic forgetting}. This is parameterized by a probability governing how often new increment data is drawn.
\end{enumerate}

\subsection{Time-Weighted Evaluation Metrics}
\label{sec:decay-metric}

In multi-episodic settings, near-future predictions are typically more operationally critical than long-term forecasts. Standard evaluations, however, weight all future samples equally, which can mask a model's responsiveness to immediate temporal shifts. To prioritize near-future accuracy, we introduce a \textit{time-weighted exponential decay} applicable to any standard metric (e.g., AUROC, MAE).

Given $N$ future evaluation windows ordered by temporal distance from the training horizon (where $i=1$ is the immediate next window) and a decay rate $\alpha \in [0, 1]$, we assign a normalized weight $W_{s, i}$ to each sample $s$ in window $i$:
\begin{equation}
    W_{s,i} = \frac{(1 - \alpha)^{i-1}}{\sum_{j=1}^{N} |S_j| (1 - \alpha)^{j-1}}
\end{equation}
where $|S_j|$ is the number of samples in window $j$. This reduces to a uniform mean when $\alpha = 0$, while $\alpha \to 1$ strictly isolates performance on the immediate next increment.

\section{Experiments}
\label{sec:experiments}

\begin{table}[t]
    \centering
    
\resizebox{\linewidth}{!}{
\begin{tabular}{@{}ll ccc cccccc@{}}
\toprule
\multirow{2}{*}{Dataset} & \multirow{2}{*}{Task} & \multicolumn{3}{c}{Original Data Splits} & \multicolumn{2}{c}{Total $N$ of $\Delta$} & \multicolumn{2}{c}{Avg. Samples} & \multicolumn{2}{c}{Span in days} \\
\cmidrule(lr){3-5} \cmidrule(lr){6-7} \cmidrule(lr){8-9} \cmidrule(lr){10-11}
 & & Train & Val & Test & $\Delta w$ & $\Delta I$ & $\Delta w$ & $\Delta I$ & $\Delta w$ & $\Delta I$ \\
\midrule
\multirow[c]{3}{*}{rel-avito} & ad-ctr & 5.1k & 1.8k & 1.8k & 5 & 4 & 1.7k & 2.3k & 4 & 6 \\
 & user-clicks & 59.5k & 21.2k & 48.0k & 5 & 4 & 20.2k & 26.9k & 4 & 6 \\
 & user-visits & 86.6k & 30.0k & 36.1k & 5 & 4 & 29.2k & 38.9k & 4 & 6 \\
\midrule
\multirow[c]{3}{*}{rel-f1} & driver-position & 7.5k & 499 & 760 & 310 & 13 & 28 & 663 & 77 & 1812 \\
 & driver-dnf & 11.4k & 566 & 702 & 475 & 13 & 27 & 998 & 48 & 1814 \\
 & driver-top3 & 1.4k & 588 & 726 & 116 & 4 & 23 & 647 & 60 & 1928 \\
\midrule
\multirow[c]{3}{*}{rel-stack} & post-votes & 2.45M & 156k & 161k & 48 & 20 & 55.5k & 137k & 91 & 225 \\
 & user-badge & 3.39M & 247k & 255k & 42 & 19 & 88.6k & 202k & 91 & 207 \\
 & user-engagement & 1.36M & 85.8k & 88.1k & 48 & 20 & 30.8k & 76.1k & 91 & 225 \\
\midrule
\multirow[c]{3}{*}{rel-trial} & site-success & 151k & 19.7k & 22.6k & 21 & 9 & 8.6k & 21.4k & 365 & 912 \\
 & study-adverse & 43.3k & 3.6k & 3.1k & 21 & 9 & 2.3k & 5.9k & 365 & 912 \\
 & study-outcome & 12.0k & 960 & 825 & 21 & 9 & 648 & 1.6k & 365 & 912 \\
\bottomrule
\end{tabular}
}
    \caption{Temporal and evaluation statistics across datasets and tasks~\cite{robinson2024relbench} used in the experiments. We report the original split sizes, alongside the total number, average sample count, and temporal span (in days) for both the evaluation windows ($\Delta w$) and increments ($\Delta I$). Large sample counts are abbreviated (k = thousands, M = millions).}
    \label{tab:task-info}
\end{table}

We evaluate our multi-episodic paradigm using 12 tasks (both regression and binary classification) defined on 4 different databases from the established RelBench~\cite{robinson2024relbench} benchmark, which currently serves as the standard for evaluating RDL methods~\cite{lachi_rgp_2025,yin_rel-moss_2026}. Through these experiments, we (i) analyze the temporal data distribution, (ii) confirm the presence of temporal concept drift, (iii) evaluate the effectiveness of knowledge transfer via incremental fine-tuning, and (iv) test the time-weighted metrics prioritizing near-future accuracy.

\subsection{Incremental Data Analysis}
Expanding upon the single-episode temporal tasks defined by future-pointing queries $Q_{\Delta w}(t_i)$ (Section~\ref{sec:rdl-tasks} and Figure~\ref{fig:temporal-tasks}), our multi-episodic paradigm incrementally evaluates models using a fixed time increment $\Delta I$ (Section~\ref{sec:mult-ep-rdl-tasks}).

Table~\ref{tab:task-info} details the temporal increments ($\Delta w$, $\Delta I$) and the original split sizes. Notably, we set the time increment $\Delta I$ based on the temporal span of the original validation split, $[t_{val}, t_{test})$, facilitating seamless comparison with the original single-episode setting~\cite{robinson2024relbench}. The table also indicates the number of available training episodes. We note that, because the final two increments correspond to the original validation and test splits and are excluded from training, tasks on the \texttt{rel-avito} dataset (due to short temporal coverage) and the \texttt{rel-f1 driver-top3} task (due to the chosen increment span) yield only two training episodes.

\begin{figure}[t]
\centering
    \begin{subfigure}{.48\textwidth}
      \centering
      \includegraphics[width=\linewidth]{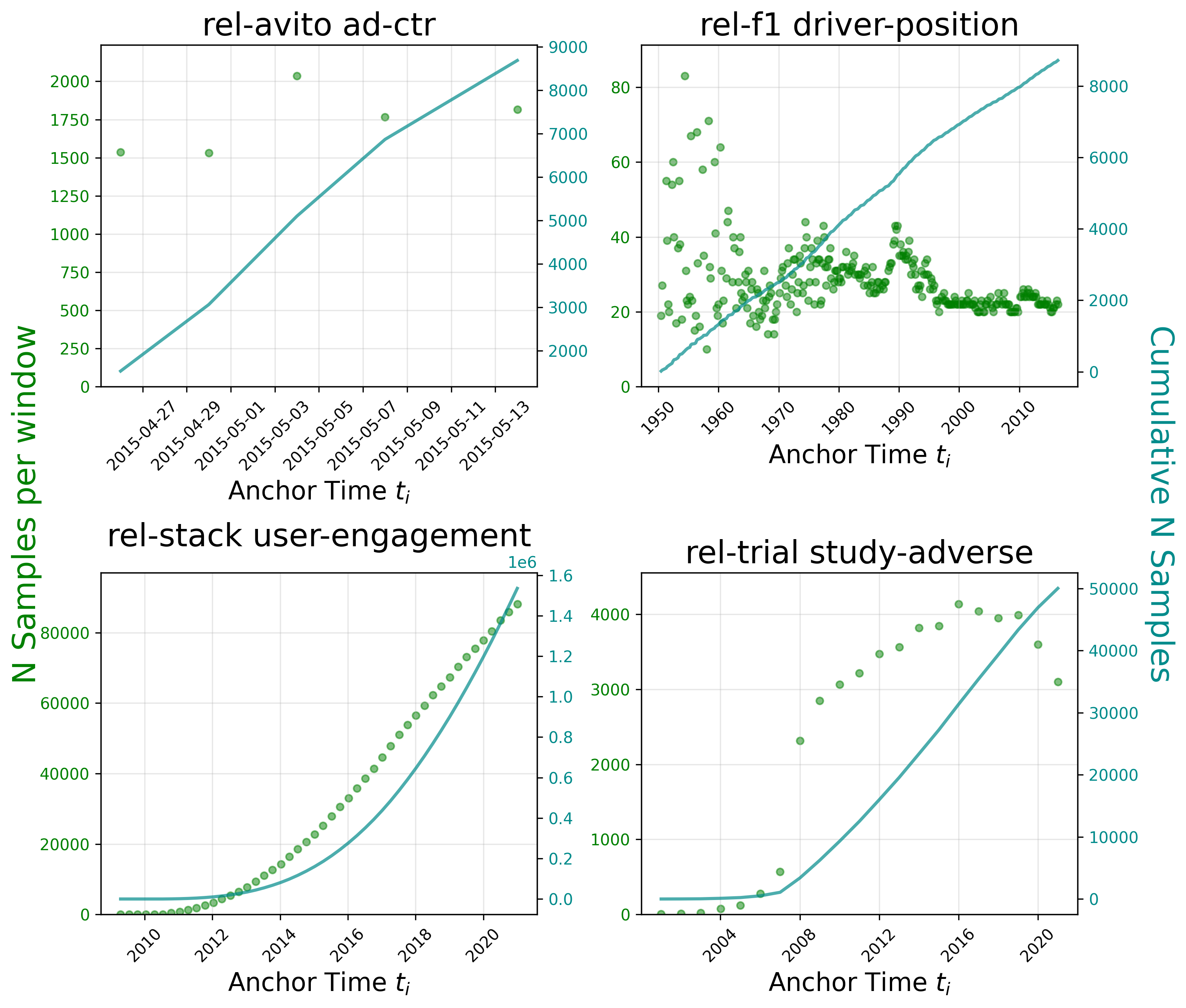}
      \caption{Number of new samples arriving in each time window $\Delta w$, together with the cumulative task size, highlighting differences in growth across successive windows in different datasets.}
      \label{fig:window-sizes-cherry}
    \end{subfigure}%
    \hfill
    \begin{subfigure}{.48\textwidth}
      \centering
      \includegraphics[width=\linewidth]{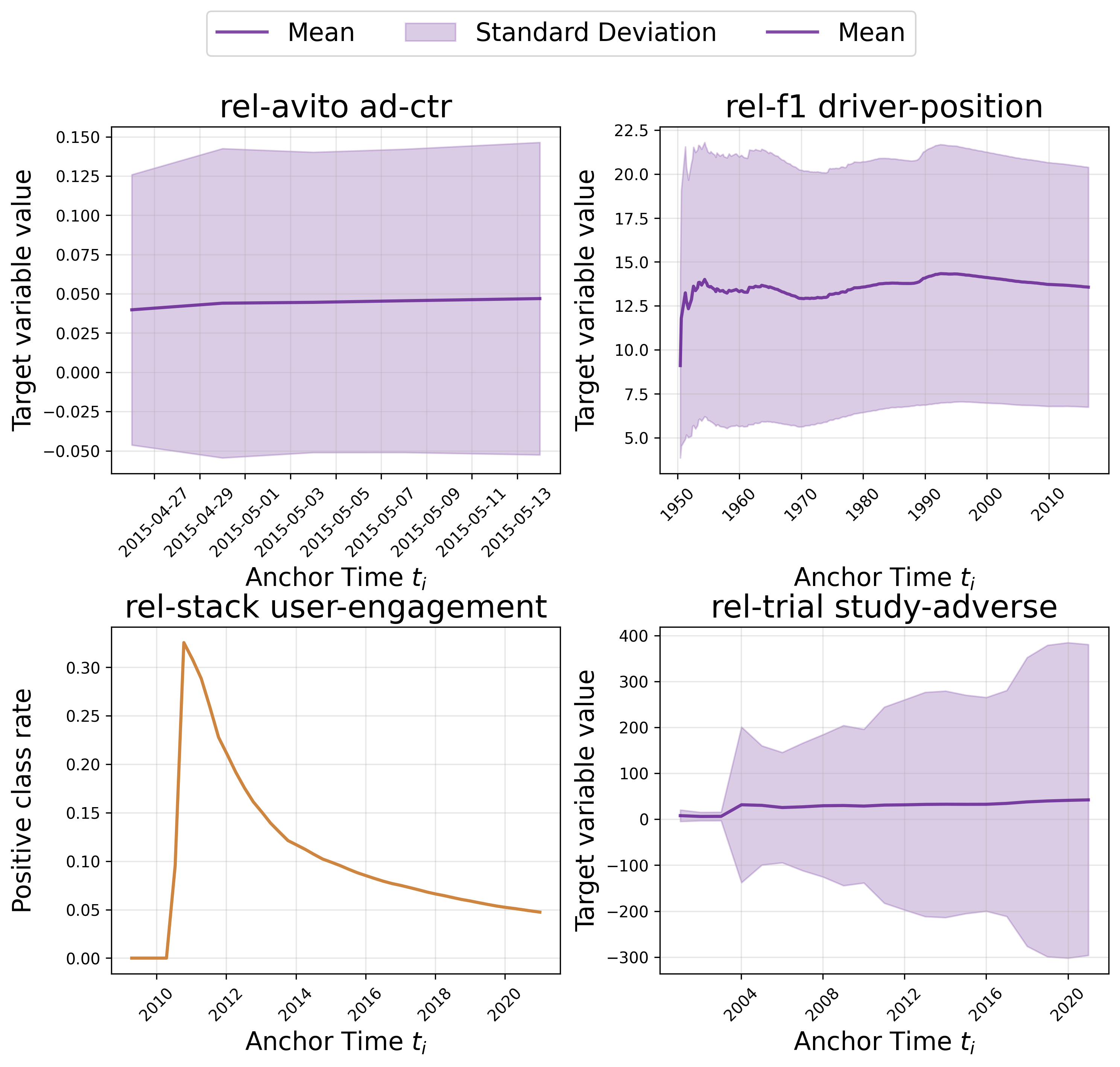}
      \caption{Cumulative value of target $y$ over the full range off the task windows $\Delta w$. Mean value with standard deviation for regression tasks, positive class rate for the binary classification tasks.}
      \label{fig:window-target-cherry}
    \end{subfigure}
\caption{Visual analysis of data distribution over time windows $\Delta w$ from anchor timestamps $t_i$
.
}
\label{fig:window-info-cherry}
\end{figure}

To detail the temporal data distribution, Figure~\ref{fig:window-info-cherry}a~\footnote{We report a subset of tasks in the main-text figures; comprehensive results are in Appendix~\ref{app:suppplementary}.} presents new sample arrivals per time window $\Delta w$ alongside cumulative task sizes. From the evolution of task sizes, we identify two principal growth patterns: roughly linear (\texttt{rel-avito ad-ctr}, \texttt{rel-f1 driver-position}) and exponential (\texttt{rel-stack user-engagement}, \texttt{rel-trial study-adverse}). Figure~\ref{fig:window-info-cherry}b illustrates the evolving cumulative target-value distributions. Notably, with the exception of \texttt{rel-avito ad-ctr} (which spans a brief interval), all remaining tasks exhibit clear evidence of temporal concept drift.

\subsection{Experimental Setup}
\label{sec:experimental-setup}

We implement our incremental evaluation and training framework using PyTorch, PyTorch Lightning, and PyTorch Geometric, building upon the RelBench~\cite{robinson2024relbench} and REDELEX~\cite{peleska_redelex_2025} libraries.

\paragraph{Model Architecture and Initialization.}
We employ a standard RDL benchmark architecture~\cite{robinson2024relbench, dwivedi_relational_2025}: tabular ResNet~\cite{Gorishniy2021RevisitingData} node encoders (via PyTorch Frame~\cite{hu2024pytorch}) paired with GraphSAGE~\cite{Hamilton2017InductiveGraphs}. The core GNN uses $2$ message-passing layers (hidden dimensionality $128$, sum aggregation), followed by task-specific prediction heads with batch normalization.

\paragraph{Graph Construction and Sampling.}
Following the RDL blueprint, databases are processed into temporal heterogeneous PK-FK graphs (Section~\ref{sec:rdl}). To handle large scales, we use mini-batch training with uniform temporal neighbor sampling, restricting context to edges and nodes valid at or prior to the anchor timestamp to strictly prevent temporal leakage (Figure~\ref{fig:temporal-tasks}). We sample a maximum of $32$ neighbors for the first GNN layer and $16$ for the second (decaying by a factor of two). For the \textit{Finetune (Upsampled)} regime, a composed data loader samples mini-batches equally ($50\%$ probability) from the newest data increment and the historical data (\figpanel{fig:continuous_RDL}{c}).

\paragraph{Training and Optimization.}
Models are optimized via Adam (learning rate $0.001$) with a batch size of $128$. During each episodic increment, models train for a maximum of $2000$ steps, artificially limited to $100$ batches per epoch to ensure frequent validation checks. All configurations are evaluated across $5$ random seeds, reporting the mean and standard deviation. 
Further compute, runtime, and orchestration details can be found in Appendix~\ref{app:compute}.

\begin{figure}[th]
    \centering
    \includegraphics[width=\linewidth]{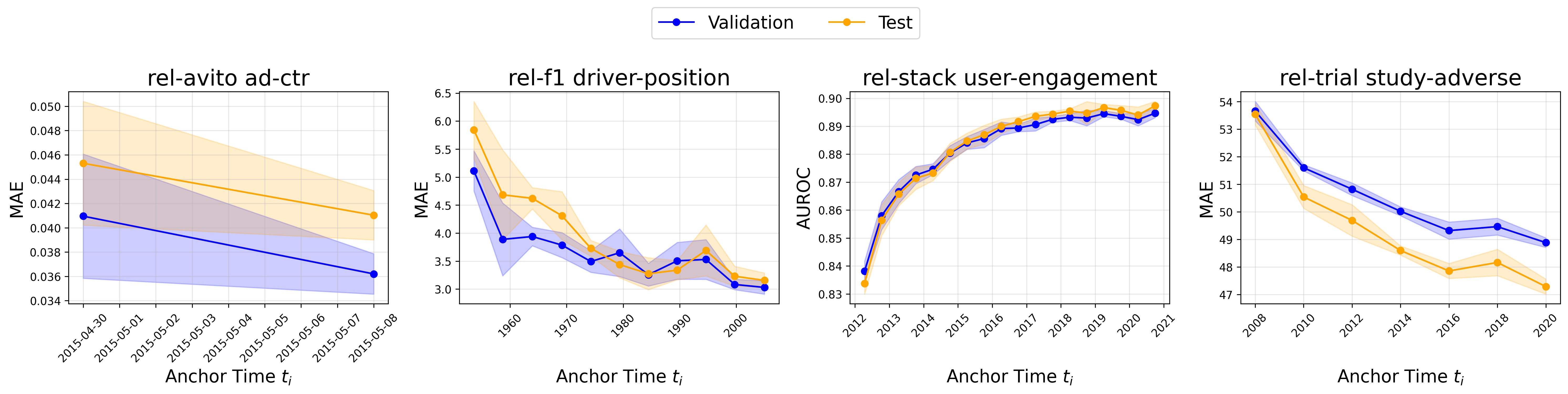}
    \caption{Performance of the \textit{From Scratch} models trained in the multi-episodic setting, reported on the original validation and test splits. Metrics are Mean Absolute Error (MAE; lower indicates better performance) for regression tasks and AUROC (higher indicates better performance) for classification.}
    \label{fig:cl-val-test-cherry}
\end{figure}

\begin{figure}[th]
\centering
    \begin{subfigure}{.48\textwidth}
        \centering
        \includegraphics[width=\linewidth]{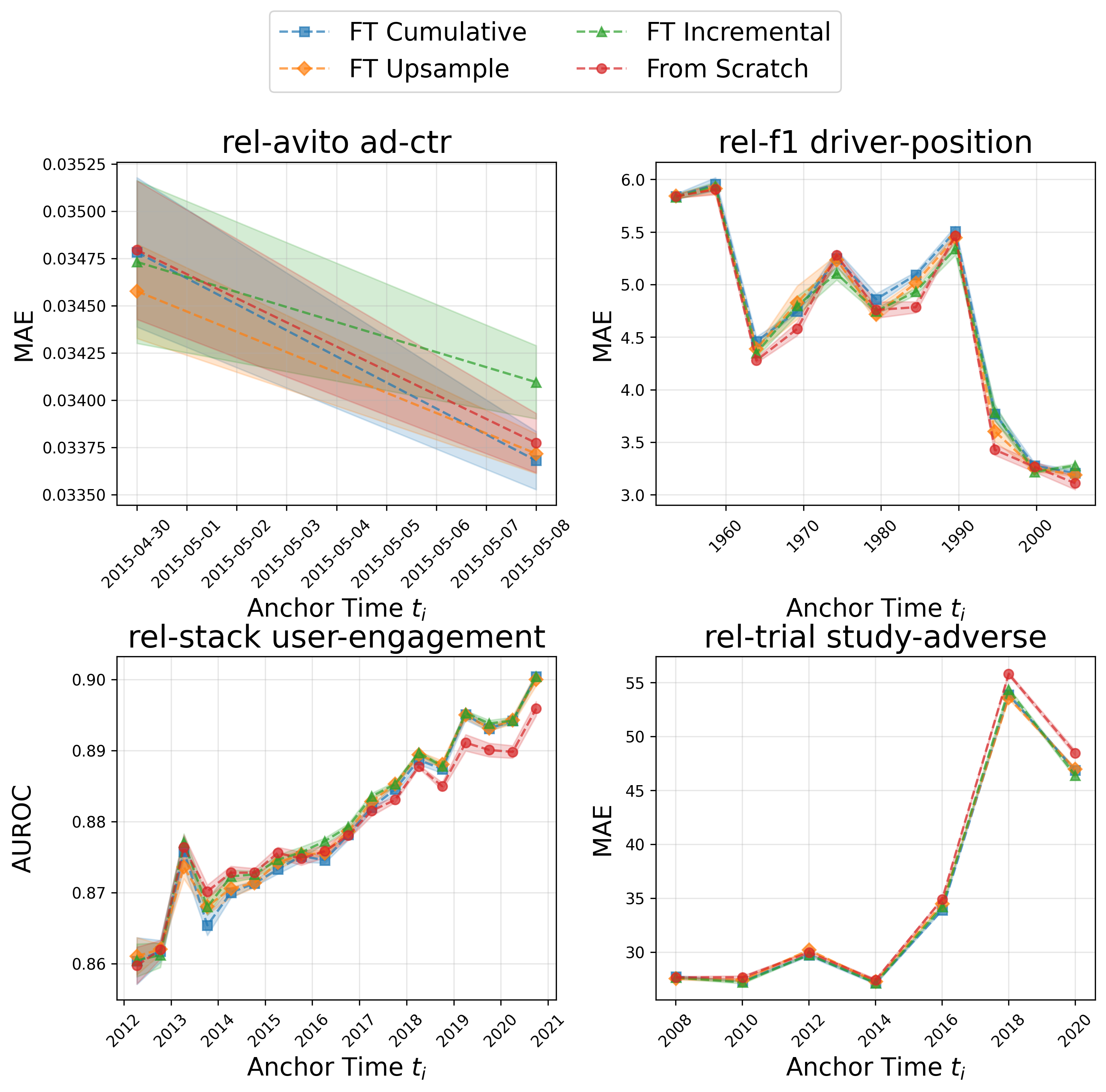}
        \caption{Top model metrics after complete training.}
        \label{fig:cl-regime-first-cherry}
    \end{subfigure}%
    \hfill
    \begin{subfigure}{.48\textwidth}
        \centering
        \includegraphics[width=\linewidth]{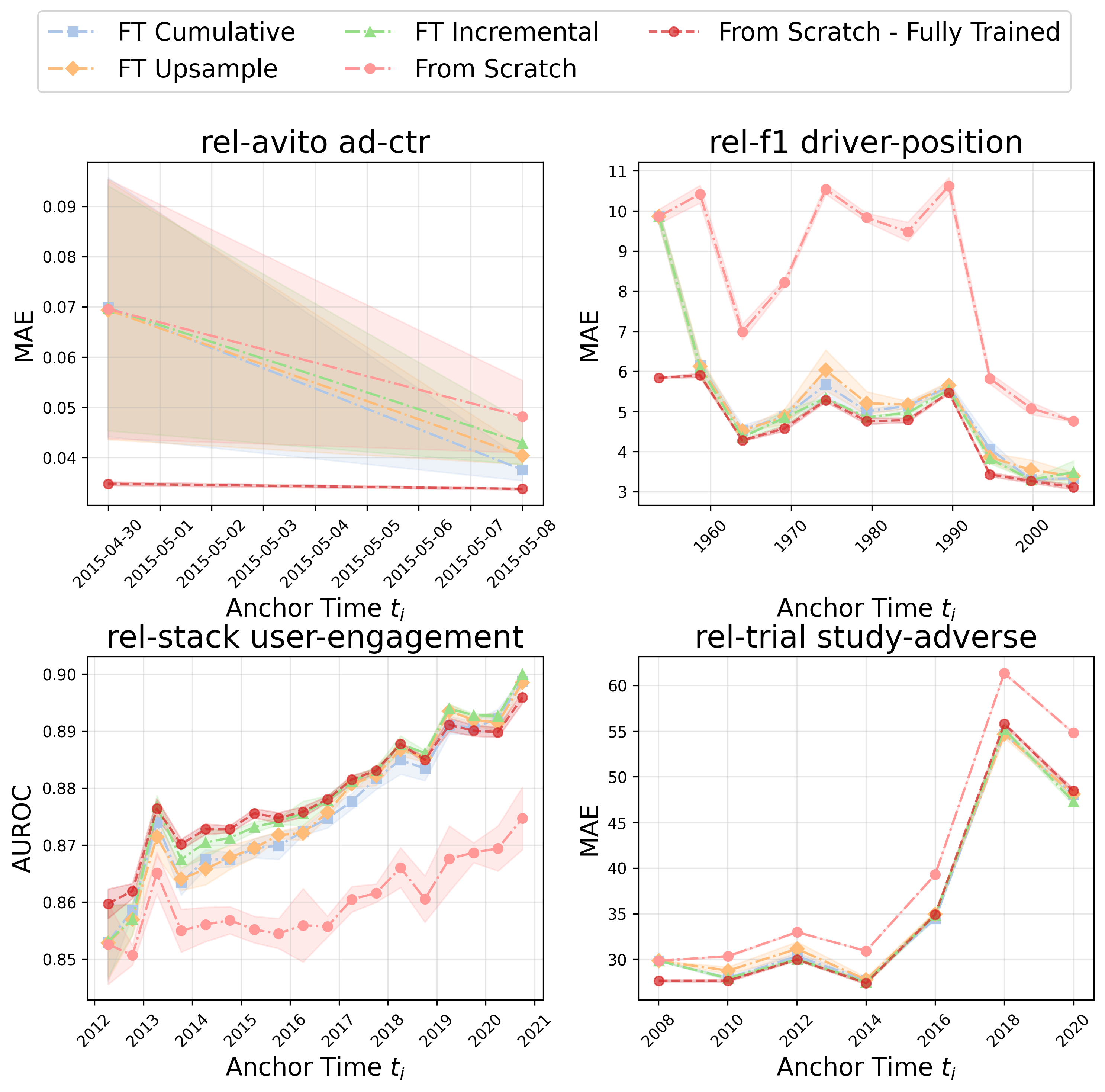}
        \caption{Model metrics after the first $100$ training steps.}
        \label{fig:cl-regime-first-cherry}
    \end{subfigure}
\caption{Performance of different model training regimes in the multi-episodic setting, reporting the validation score for the subsequent increment. Metrics are, again, MAE (lower indicates better performance) and AUROC (higher indicates better performance).}
\label{fig:cl-regime-cherry}
\end{figure}

\subsection{Investigating Temporal Concept Drift}
To confirm that the observed distribution shifts constitute temporal concept drift, we incrementally train baseline \textit{From Scratch} models and assess them on the original validation and test splits. Concretely, given anchor times $T_{\Delta I} = \{t_i \mid t_i = t_{i-1} + \Delta I, t_i \leq t_{val}\}$,\footnote{$t_{val}$ without a superscript refers to the original (non-incremental) dataset split boundary} each model $M_i$ is trained up to $t_i$, with the best checkpoint selected via validation on the subsequent increment (up to $t_{i+1}$).

Figure~\ref{fig:cl-val-test-cherry} then substantiates the presence of concept drift through a general trend of performance degradation over time across tasks. A minor exception to that trend is \texttt{rel-f1 driver-position}, which exhibits a substantial drop in performance, likely attributable to its relatively small dataset size. Nevertheless, the overall empirical validation of the drift confirms the original motivation for the RDL focus on temporal dynamics.

\subsection{Evaluating Fine-tuning Regimes}

Our multi-episodic setting enables RDL methods that exploit incrementally expanding data. Figure~\ref{fig:cl-regime-cherry}a compares the top performances of our three fine-tuning regimes (Section~\ref{sec:incr-train-regime}) against a \textit{From Scratch} baseline. The fine-tuning regimes consistently match or outperform the baseline, an effect most pronounced in the \texttt{rel-stack user-engagement} and \texttt{rel-trial study-adverse} tasks.

To further illustrate knowledge transfer capabilities, Figure~\ref{fig:cl-regime-first-cherry} shows model performance after just $100$ training steps, resembling a \textit{Few-Shot Learning} scenario. Notably, pretrained models quickly achieve performance on par with or superior to a fully trained \textit{from scratch} baseline, underscoring the computational efficiency and predictive potential of continuously fine-tuned models.

\begin{table}[th]
    \centering
    \small{
\begin{tabular}{lllrrrr}
    \toprule
    \multirow{2}{*}{Dataset} & \multirow{2}{*}{Task} & \multirow{2}{*}{Decay} & \multicolumn{4}{c}{Training Regime - MAE $\downarrow$}  \\
    \cmidrule(lr){4-7} 
 & & & FT Cuml. & FT Incr. & FT Upspl. & Scratch  \\
    \midrule
    \multirow[c]{3}{*}{rel-avito} & \multirow[c]{3}{*}{ad-ctr} & 0.0 & $\mathbf{0.037}$ & $\mathbf{0.037}$ & $0.039$ & $0.039$ \\
     &  & 0.3 & $\mathbf{0.037}$ & $\mathbf{0.037}$ & $0.039$ & $0.038$ \\
     &  & 0.9 & $\mathbf{0.035}$ & $0.036$ & $0.037$ & $0.037$ \\
    \midrule
    \multirow[c]{3}{*}{rel-f1} & \multirow[c]{3}{*}{driver-position} & 0.0 & $3.256$ & $3.408$ & $3.278$ & $\mathbf{3.106}$ \\
     &  & 0.3 & $2.858$ & $2.922$ & $3.058$ & $\mathbf{2.827}$ \\
     &  & 0.9 & $3.466$ & $\mathbf{3.293}$ & $3.569$ & $3.358$ \\
    \midrule
    \multirow[c]{3}{*}{rel-stack} & \multirow[c]{3}{*}{post-votes} & 0.0 & $\mathbf{0.061}$ & $0.062$ & $0.062$ & $0.066$ \\
     &  & 0.3 & $\mathbf{0.061}$ & $0.062$ & $0.062$ & $0.065$ \\
     &  & 0.9 & $\mathbf{0.059}$ & $0.060$ & $0.060$ & $0.063$ \\
    \midrule
    \multirow[c]{6}{*}{rel-trial} & \multirow[c]{3}{*}{site-success} & 0.0 & $0.428$ & $0.395$ & $0.419$ & $\mathbf{0.393}$ \\
     &  & 0.3 & $0.431$ & $0.396$ & $0.420$ & $\mathbf{0.392}$ \\
     &  & 0.9 & $0.440$ & $0.399$ & $0.421$ & $\mathbf{0.389}$ \\
    \cline{2-7}
     & \multirow[c]{3}{*}{study-adverse} & 0.0 & $46.369$ & $\mathbf{45.996}$ & $46.545$ & $48.143$ \\
     &  & 0.3 & $46.565$ & $\mathbf{46.163}$ & $46.819$ & $48.281$ \\
     &  & 0.9 & $47.239$ & $\mathbf{46.735}$ & $47.760$ & $48.755$ \\
    \bottomrule
\end{tabular}
}

    \caption{Performance of individual training regimes, trained on the complete training split, assessed using the temporally decayed MAE over the combined original validation and test splits.}
    \label{tab:decay-reg}
\end{table}

\subsection{Decayed Evaluation Metrics}
To account for the operational prioritization of near-future predictions, Table~\ref{tab:decay-reg} applies our temporally decayed metrics (Section~\ref{sec:decay-metric}) to the regression tasks. A decay of $0.0$ yields a standard uniform average, while $0.9$ places almost all weight on the immediate next window. Notably, except for \texttt{driver-position}, varying the decay factor preserves the optimal training regime rankings. Nevertheless, absolute metric values shift in distinct patterns: \texttt{post-votes} and \texttt{ad-ctr} exhibit minimal variation, \texttt{study-adverse} shows a consistent performance drop, and \texttt{driver-position} improves at $\alpha=0.3$---a variation we suspect reflects differing rates of localized temporal volatility across datasets, with a detailed analysis left for future work.

\section{Conclusion}
\label{sec:conclusion}

In this work, we addressed a major shortcoming in standard Relational Deep Learning (RDL) evaluation: the reliance on static dataset snapshots, which fundamentally misaligns with the continuous, dynamic evolution of real-world databases. By formalizing an incremental, multi-episodic evaluation and training paradigm, we provide a framework that accurately reflects the real-world lifecycle of database growth and model deployment.

Our empirical analysis across large-scale relational datasets establishes several core findings. First, we validated the widespread prevalence of temporal concept drift in standard RDL tasks, highlighting why snapshot-based testing is insufficient. Furthermore, we showed that incremental fine-tuning strategies successfully mitigate this degradation. Transferring knowledge from previous training episodes consistently matched or exceeded the performance of expensive from-scratch retraining, often adapting within just a few optimization steps. Finally, introducing a time-weighted exponential decay metric allowed for a more realistic assessment of a model's utility by properly prioritizing near-future predictive accuracy. 

Altogether, the proposed paradigm and associated training regimes lay the groundwork for transitioning RDL research from isolated benchmarks toward practical, continually learning systems.

\paragraph{Limitations.}
\label{sec:limitations}

While our experimental setup currently focuses on append-only database growth, real-world systems often involve data updates and deletions (CRUD operations)~\cite{Codd1970}. Although our multi-episodic paradigm can in principle accommodate these mutations---such as by dynamically updating historical node states or utilizing temporal edges~\cite{kazemi2020representation}---doing so introduces complex attribute dependencies that require further formal investigation. Furthermore, while we establish foundational fine-tuning regimes, mitigating catastrophic forgetting over extended horizons remains a challenge~\cite{zhang2022continual}. Although the incremental fine-tuning offers a highly scalable alternative as cumulative data grows, optimally balancing this computational efficiency with long-term knowledge retention may require integrating more advanced continual learning techniques, such as dynamic replay buffers or weight regularization~\cite{parisi2019continual}. Finally, extending this continuous evaluation to emerging monolithic RDL foundation models~\cite{wehrstein_towards_2025,hudovernik_kumorfm-2_2026,ranjan_relational_2025,wang_griffin_2025} remains an open challenge to fully understand their specific resilience to temporal concept drift.


\begin{ack}

\end{ack}

\bibliographystyle{abbrv}
\bibliography{references}
\newpage


\appendix

\section{Technical appendices and supplementary material}
\label{app:suppplementary}

This appendix provides additional implementation details, full data-distribution analyses, and extended experimental results that complement the main text.

\subsection{Detailed Experimental Setup and Compute Resources}
\label{app:compute}

We expand the experimental overview from Section~\ref{sec:experimental-setup} by summarizing the implementation stack and the main architectural, data-loading, and training choices. All models are implemented in PyTorch using PyTorch Lightning and PyTorch Geometric, and integrate with the RelBench~\cite{robinson2024relbench} and REDELEX~\cite{peleska_redelex_2025} libraries.

\paragraph{Model Architecture and Initialization.}
As formalized in the source scripts, the core architecture relies on a \texttt{HeterogeneousSAGE} module. The relational databases are processed into temporal heterogeneous graphs following the primary key-foreign key (PK-FK) relationships. To initialize the node features from the raw database attributes, we utilize \texttt{HeteroEncoder} built upon PyTorch Frame~\cite{hu2024pytorch}, leveraging tabular architectures such as ResNet~\cite{Gorishniy2021RevisitingData} to produce initial node embedding matrices of dimensionality 128.

Crucially, temporal dynamics are injected via a \texttt{HeteroTemporalEncoder}. For a given prediction instance, this module computes relative time encodings based on the \textit{anchor time} of the target entity (Section~\ref{sec:rdl-tasks}). These temporal embeddings are directly added to the initial node features before being passed to the GNN. The graph message passing is handled by a 2-layer \texttt{HeteroGraphSAGE} using sum aggregation. Finally, a single-layer MLP with batch normalization serves as the task-specific prediction head. All components are trained end-to-end to jointly learn the tabular representations and the graph structure.

\paragraph{Graph Construction and Sampling.}
Given the large scale of the datasets, we utilize mini-batch training with a temporal neighbor sampling strategy. Specifically, we sample a maximum of $32$ neighbors for the first GNN layer and $16$ neighbors for the second layer (decaying by a factor of two per layer). To strictly prevent temporal leakage, we use a uniform temporal sampling strategy that only considers edges and nodes valid at or prior to the anchor timestamp (Section~\ref{sec:rdl-tasks}). For the \textit{Finetune (Upsampled)} training regime (Section~\ref{sec:incr-train-regime}), we construct a specialized data loader that samples mini-batches from the newest data increment and the historical data with an equal $50\%$ probability.

\paragraph{Training and Optimization.}
The models are optimized using the Adam optimizer with an initial learning rate of $0.001$. We use a uniform batch size of $128$ for all tasks. During each episodic increment (Section~\ref{sec:incr-eval}), models are trained for a maximum of $2000$ optimization steps. We artificially limit the training epochs to $100$ batches per epoch to ensure frequent validation checks and early stopping.

All experiments were orchestrated using Ray~\cite{moritz2018ray} to manage the distributed execution of the incremental training regimes (Section~\ref{sec:incr-train-regime}) across different time splits. Trials were automatically scheduled on available GPU resources, most commonly on the \texttt{Tesla V100-SXM2-32GB}. To ensure computational feasibility and simulate realistic production constraints, a strict maximum time limit of 2 hours was enforced for each individual incremental training step. Model checkpointing was handled by PyTorch Lightning, saving the best model weights based on the validation metric at each epoch.

\subsection{Data Analysis}

\begin{figure}[H]
    \centering
    \includegraphics[width=0.95\linewidth]{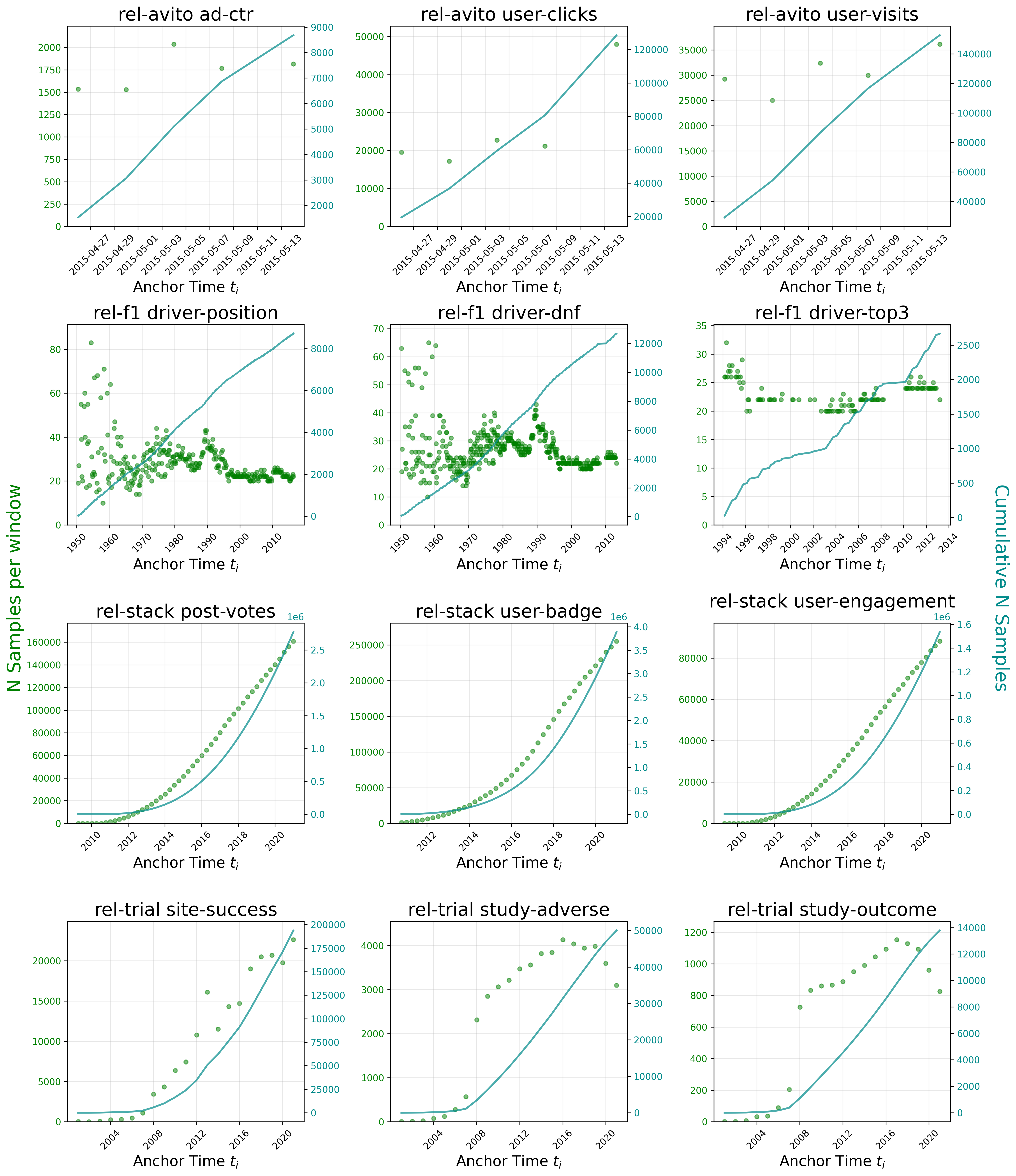}
    \caption{Number of new samples from individual time windows $\Delta w$ and the cumulative size of the task over time.}
    \label{fig:window-sizes-full}
\end{figure}

\begin{figure}[H]
    \centering
    \includegraphics[width=0.95\linewidth]{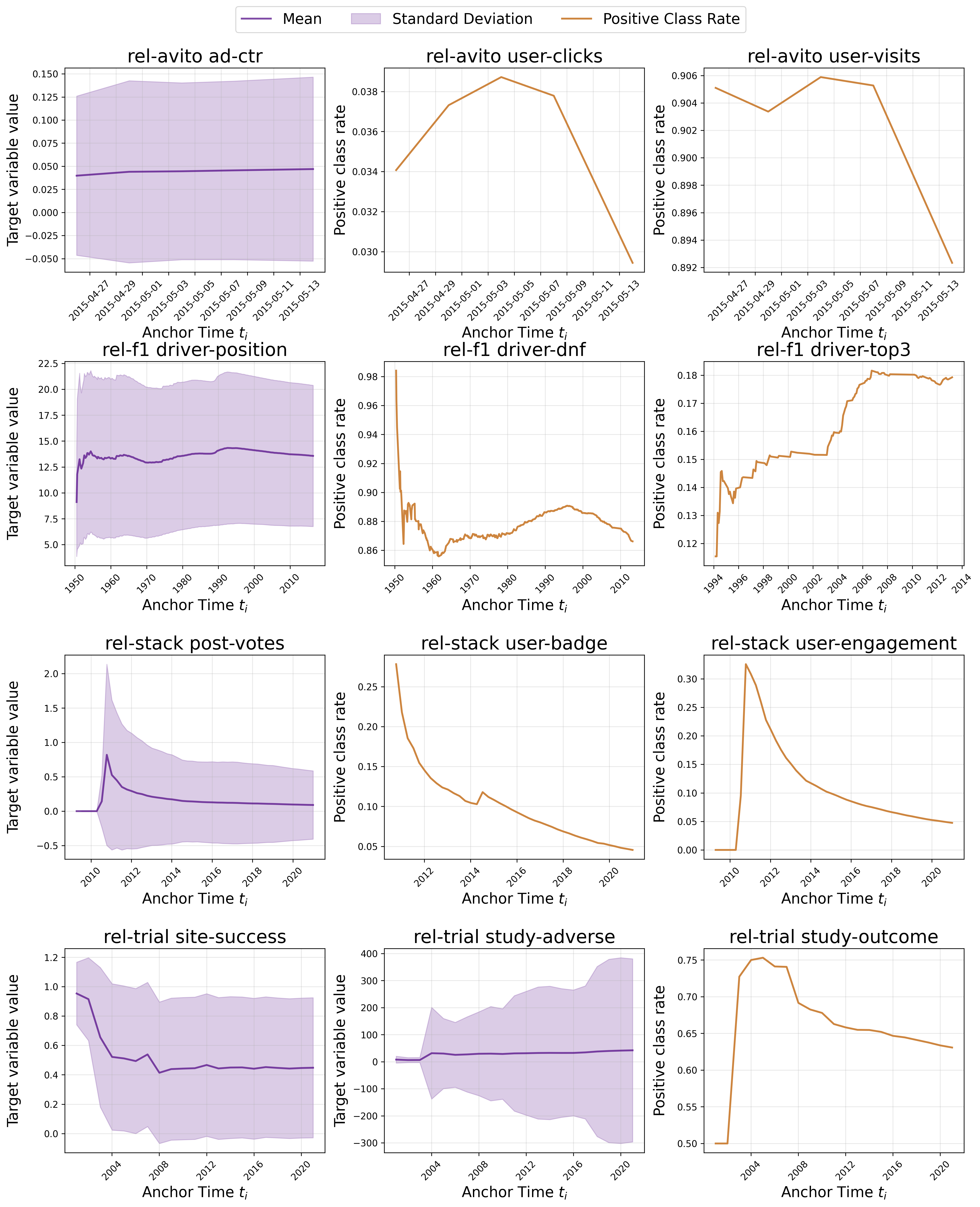}
    \caption{Cumulative value of target $y$ over the full range off the task windows $\Delta w$. Mean value with standard deviation for regression tasks, positive class rate for the binary classification tasks.}
    \label{fig:window-target-full}
\end{figure}

Figures~\ref{fig:window-sizes-full} and \ref{fig:window-target-full} present the full temporal distribution analysis for all 12 evaluated tasks. Whereas the main text focused on selected illustrative cases, these complete results confirm that the same qualitative patterns described there also hold more generally. In Figure~\ref{fig:window-sizes-full}, we see that tasks defined on a shared dataset exhibit similar distributions of samples across time windows (Section~\ref{sec:mult-ep-rdl-tasks}), including the categorization into the `linear' and `exponential' categories. In contrast, Figure~\ref{fig:window-target-full} shows that the target value distributions vary considerably between tasks, with pronounced distributional shifts that are particularly evident for the tasks derived from the \texttt{rel-f1} dataset.

\subsection{Extended Results}

This section reports the complete set of performance metrics for all evaluated tasks, extending the illustrative examples provided in the main text.

\begin{figure}[H]
    \centering
    \includegraphics[width=0.95\linewidth]{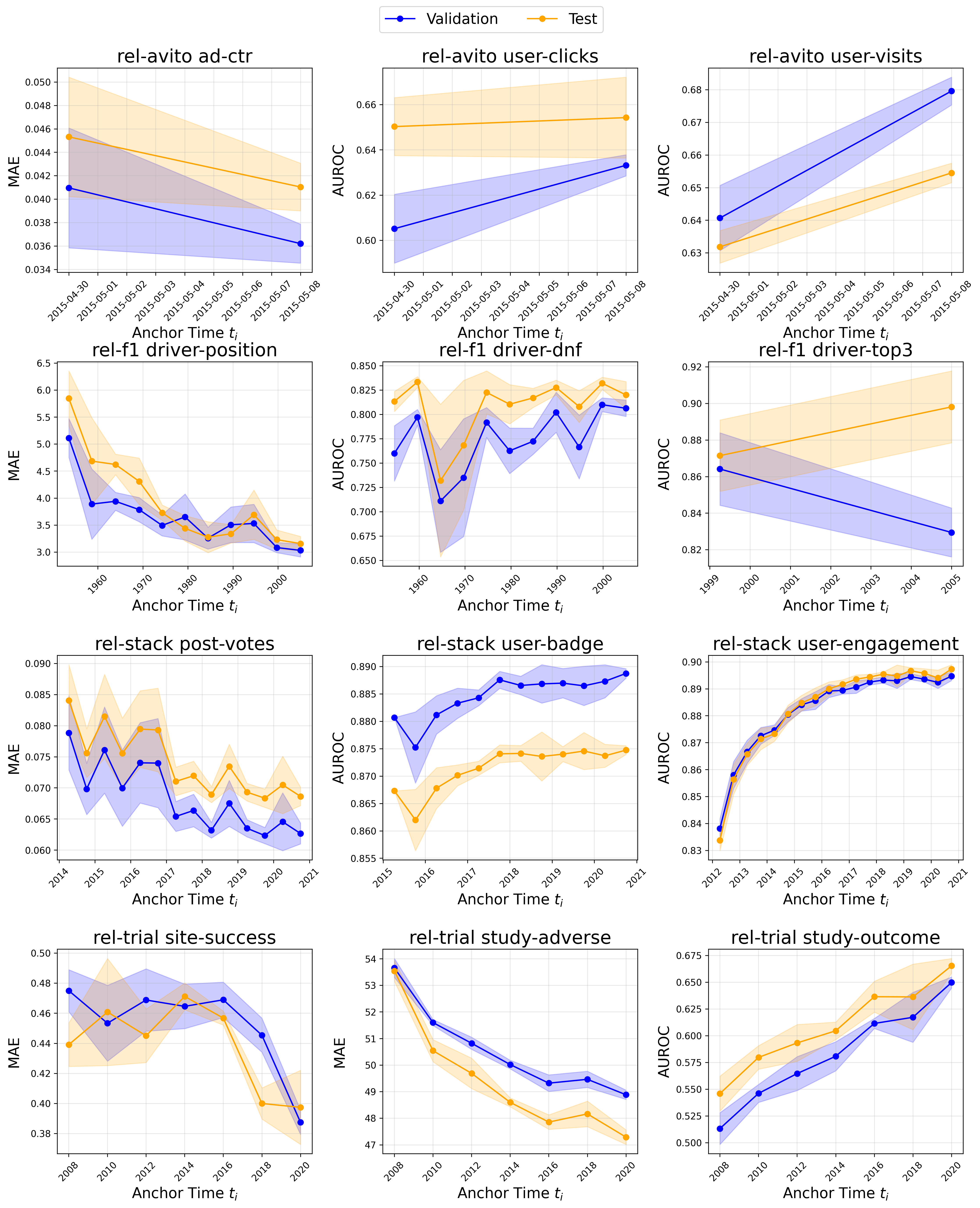}
    \caption{Performance of the \textit{From Scratch} models trained in the multi-episodic setting, reported on the original validation and test splits. Metrics are Mean Absolute Error (MAE; lower indicates better performance) for regression tasks and AUROC (higher indicates better performance) for binary classification.}
    \label{fig:cl-val-test-full}
\end{figure}

Figure~\ref{fig:cl-val-test-full} presents the performance of the \textit{From Scratch} baseline (Section~\ref{sec:incr-train-regime}) across all increments (Section~\ref{sec:incr-eval}). With the exception of \texttt{rel-f1 driver-dnf} and \texttt{rel-trial site-success}, all tasks display a clear performance improvement for episodes nearer to the validation horizon, underscoring the susceptibility of static models to temporal concept drift.

\begin{figure}[H]
    \centering
    \includegraphics[width=0.95\linewidth]{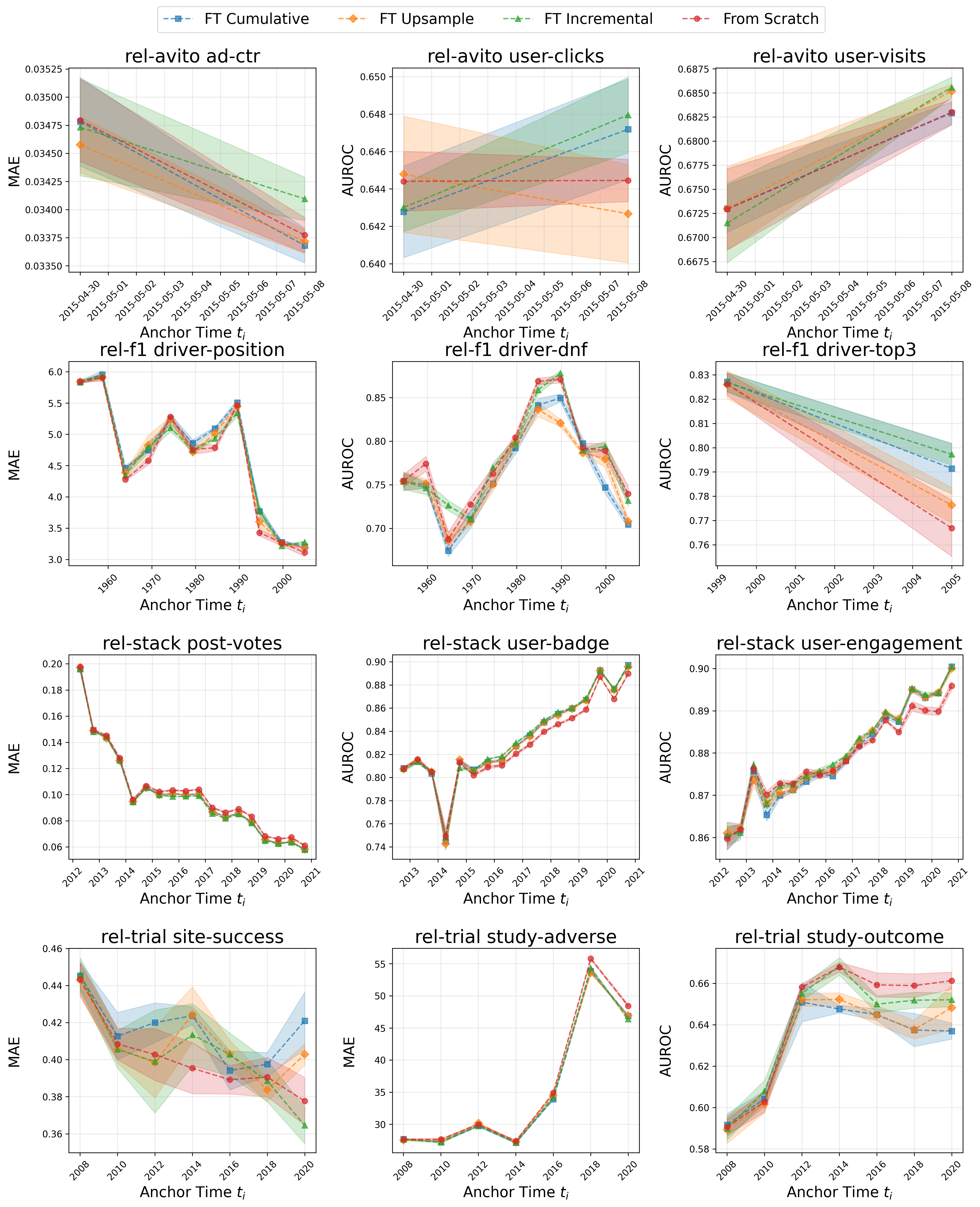}
    \caption{Top performance of different model training regimes in the multi-episodic setting, reporting the validation score for the subsequent increment. Metrics are Mean Absolute Error (MAE; lower indicates better performance) for regression tasks and AUROC (higher indicates better performance) for binary classification.}
    \label{fig:cl-regime-best-full}
\end{figure}

\begin{figure}[H]
    \centering
    \includegraphics[width=0.95\linewidth]{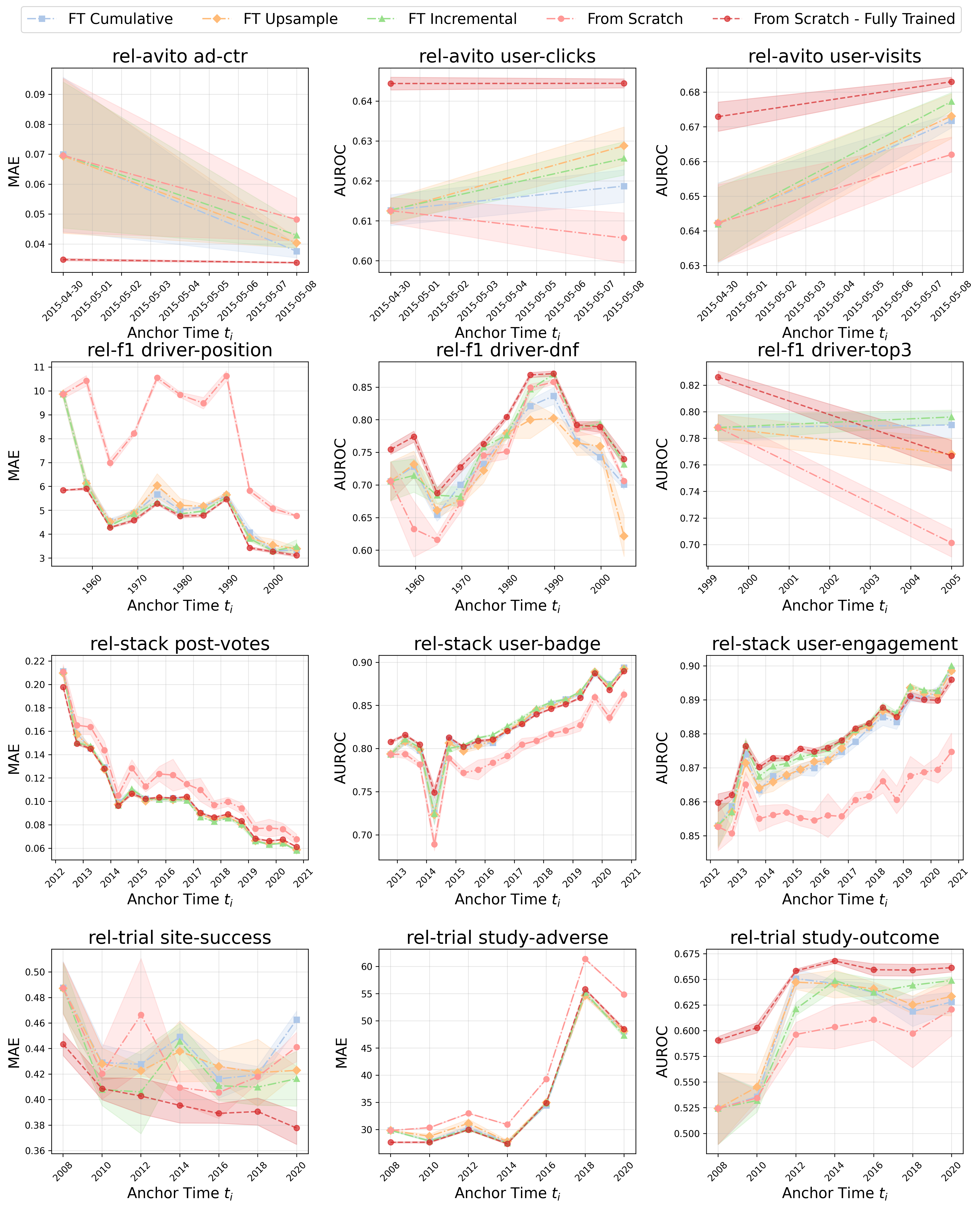}
    \caption{Performance of different model training regimes in the multi-episodic setting after first $100$ training steps, reporting the validation score for the subsequent increment. Metrics are Mean Absolute Error (MAE; lower indicates better performance) for regression tasks and AUROC (higher indicates better performance) for binary classification.}
    \label{fig:cl-regime-first-full}
\end{figure}

Figures~\ref{fig:cl-regime-best-full} and \ref{fig:cl-regime-first-full} contrast the different proposed incremental training regimes (Section~\ref{sec:incr-train-regime}). The results consistently indicate that our fine-tuning strategies substantially enhance performance in the few-shot scenario shown in Figure~\ref{fig:cl-regime-first-full}, highlighting the potential of knowledge transfer between the episodes, and frequently also improve fully trained models, with the effect being especially pronounced for tasks defined on the \texttt{rel-stack} database.

Tables~\ref{tab:decay-reg-std} and \ref{tab:decay-cls-std} broaden the time-weighted evaluation from the main text (Section~\ref{sec:decay-metric}) by reporting standard deviations over 5 independent runs for regression and classification tasks, respectively. Notably, Table~\ref{tab:decay-cls-std} highlights the consistently superior performance of the fine-tuning regimes, with the sole exception of \texttt{rel-trial study-outcome}, which remains challenging, as traditional tabular models can outperform RDL~\cite{robinson2024relbench} on this task.

\begin{table}[H]
    \centering
    \resizebox{\linewidth}{!}{
\begin{tabular}{lllrrrr}
    \toprule
    \multirow{2}{*}{Dataset} & \multirow{2}{*}{Task} & \multirow{2}{*}{Decay} & \multicolumn{4}{c}{Training Regime}  \\
    \cmidrule(lr){4-7} 
 & & & FT Cuml. & FT Incr. & FT Upspl. & Scratch  \\
    \midrule
    \multirow[c]{3}{*}{rel-avito} & \multirow[c]{3}{*}{ad-ctr} & 0.0 & $\mathbf{0.037} \pm 0.001$ & $\mathbf{0.037} \pm 0.001$ & $0.039 \pm 0.002$ & $0.039 \pm 0.002$ \\
     &  & 0.3 & $\mathbf{0.037} \pm 0.001$ & $\mathbf{0.037} \pm 0.001$ & $0.039 \pm 0.002$ & $0.038 \pm 0.002$ \\
     &  & 0.9 & $\mathbf{0.035} \pm 0.001$ & $0.036 \pm 0.001$ & $0.037 \pm 0.002$ & $0.037 \pm 0.002$ \\
    \midrule
    \multirow[c]{3}{*}{rel-f1} & \multirow[c]{3}{*}{driver-position} & 0.0 & $3.256 \pm 0.056$ & $3.408 \pm 0.053$ & $3.278 \pm 0.104$ & $\mathbf{3.106} \pm 0.125$ \\
     &  & 0.3 & $2.858 \pm 0.055$ & $2.922 \pm 0.021$ & $3.058 \pm 0.153$ & $\mathbf{2.827} \pm 0.116$ \\
     &  & 0.9 & $3.466 \pm 0.135$ & $\mathbf{3.293} \pm 0.107$ & $3.569 \pm 0.316$ & $3.358 \pm 0.247$ \\
    \midrule
    \multirow[c]{3}{*}{rel-stack} & \multirow[c]{3}{*}{post-votes} & 0.0 & $\mathbf{0.061} \pm 0.000$ & $0.062 \pm 0.001$ & $0.062 \pm 0.001$ & $0.066 \pm 0.002$ \\
     &  & 0.3 & $\mathbf{0.061} \pm 0.001$ & $0.062 \pm 0.001$ & $0.062 \pm 0.001$ & $0.065 \pm 0.002$ \\
     &  & 0.9 & $\mathbf{0.059} \pm 0.001$ & $0.060 \pm 0.001$ & $0.060 \pm 0.001$ & $0.063 \pm 0.002$ \\
    \midrule
    \multirow[c]{6}{*}{rel-trial} & \multirow[c]{3}{*}{site-success} & 0.0 & $0.428 \pm 0.017$ & $0.395 \pm 0.003$ & $0.419 \pm 0.008$ & $\mathbf{0.393} \pm 0.013$ \\
     &  & 0.3 & $0.431 \pm 0.016$ & $0.396 \pm 0.003$ & $0.420 \pm 0.006$ & $\mathbf{0.392} \pm 0.011$ \\
     &  & 0.9 & $0.440 \pm 0.014$ & $0.399 \pm 0.005$ & $0.421 \pm 0.006$ & $\mathbf{0.389} \pm 0.007$ \\
    \cline{2-7}
     & \multirow[c]{3}{*}{study-adverse} & 0.0 & $46.369 \pm 0.104$ & $\mathbf{45.996} \pm 0.321$ & $46.545 \pm 0.304$ & $48.143 \pm 0.183$ \\
     &  & 0.3 & $46.565 \pm 0.092$ & $\mathbf{46.163} \pm 0.300$ & $46.819 \pm 0.299$ & $48.281 \pm 0.175$ \\
     &  & 0.9 & $47.239 \pm 0.120$ & $\mathbf{46.735} \pm 0.267$ & $47.760 \pm 0.327$ & $48.755 \pm 0.170$ \\
    \bottomrule
\end{tabular}
}
    \caption{Performance of individual training regimes on regression tasks, trained on the complete training split, assessed
using temporally decayed Mean Absolute Error with standard deviations (lower indicates better performance) over the
combined original validation and test splits.}
    \label{tab:decay-reg-std}
\end{table}

\begin{table}[H]
    \centering
    \resizebox{\linewidth}{!}{
\begin{tabular}{lllrrrr}
    \toprule
    \multirow{2}{*}{Dataset} & \multirow{2}{*}{Task} & \multirow{2}{*}{Decay} & \multicolumn{4}{c}{Training Regime}  \\
    \cmidrule(lr){4-7} 
    & & & FT Cuml. & FT Incr. & FT Upspl. & Scratch  \\
    \midrule
    \multirow[c]{6}{*}{rel-avito} & \multirow[c]{3}{*}{user-clicks} & 0.0 & $\mathbf{0.659} \pm 0.005$ & $\mathbf{0.659} \pm 0.006$ & $0.656 \pm 0.011$ & $0.648 \pm 0.013$ \\
     &  & 0.3 & $\mathbf{0.657} \pm 0.004$ & $\mathbf{0.657} \pm 0.006$ & $0.654 \pm 0.010$ & $0.646 \pm 0.012$ \\
     &  & 0.9 & $0.645 \pm 0.002$ & $\mathbf{0.646} \pm 0.004$ & $0.641 \pm 0.007$ & $0.637 \pm 0.006$ \\
    \cline{2-7}
     & \multirow[c]{3}{*}{user-visits} & 0.0 & $0.665 \pm 0.003$ & $\mathbf{0.666} \pm 0.002$ & $\mathbf{0.666} \pm 0.001$ & $\mathbf{0.666} \pm 0.003$ \\
     &  & 0.3 & $0.667 \pm 0.003$ & $\mathbf{0.669} \pm 0.002$ & $\mathbf{0.669} \pm 0.001$ & $0.668 \pm 0.004$ \\
     &  & 0.9 & $0.676 \pm 0.003$ & $\mathbf{0.678} \pm 0.002$ & $\mathbf{0.678} \pm 0.001$ & $0.677 \pm 0.004$ \\
    \midrule
    \multirow[c]{6}{*}{rel-f1} & \multirow[c]{3}{*}{driver-dnf} & 0.0 & $0.763 \pm 0.016$ & $\mathbf{0.818} \pm 0.001$ & $0.776 \pm 0.008$ & $0.814 \pm 0.011$ \\
     &  & 0.3 & $0.737 \pm 0.034$ & $\mathbf{0.759} \pm 0.015$ & $0.745 \pm 0.024$ & $0.755 \pm 0.016$ \\
     &  & 0.9 & $0.695 \pm 0.074$ & $\mathbf{0.716} \pm 0.004$ & $0.687 \pm 0.053$ & $0.670 \pm 0.025$ \\
    \cline{2-7}
     & \multirow[c]{3}{*}{driver-top3} & 0.0 & $0.892 \pm 0.006$ & $\mathbf{0.898} \pm 0.003$ & $0.841 \pm 0.009$ & $0.867 \pm 0.016$ \\
     &  & 0.3 & $\mathbf{0.779} \pm 0.007$ & $\mathbf{0.779} \pm 0.009$ & $0.725 \pm 0.029$ & $0.754 \pm 0.020$ \\
     &  & 0.9 & $\mathbf{0.689} \pm 0.018$ & $0.671 \pm 0.016$ & $0.659 \pm 0.053$ & $0.653 \pm 0.048$ \\
    \midrule
    \multirow[c]{6}{*}{rel-stack} & \multirow[c]{3}{*}{user-badge} & 0.0 & $0.889 \pm 0.000$ & $\mathbf{0.890} \pm 0.001$ & $0.889 \pm 0.001$ & $0.882 \pm 0.000$ \\
     &  & 0.3 & $0.890 \pm 0.000$ & $\mathbf{0.891} \pm 0.001$ & $0.890 \pm 0.001$ & $0.883 \pm 0.001$ \\
     &  & 0.9 & $\mathbf{0.894} \pm 0.000$ & $\mathbf{0.894} \pm 0.001$ & $0.893 \pm 0.001$ & $0.887 \pm 0.001$ \\
    \cline{2-7}
     & \multirow[c]{3}{*}{user-engagement} & 0.0 & $\mathbf{0.902} \pm 0.001$ & $0.901 \pm 0.001$ & $0.901 \pm 0.001$ & $0.896 \pm 0.001$ \\
     &  & 0.3 & $\mathbf{0.902} \pm 0.001$ & $0.901 \pm 0.001$ & $0.901 \pm 0.001$ & $0.896 \pm 0.001$ \\
     &  & 0.9 & $\mathbf{0.901} \pm 0.001$ & $0.900 \pm 0.001$ & $0.899 \pm 0.001$ & $0.895 \pm 0.002$ \\
    \midrule
    \multirow[c]{3}{*}{rel-trial} & \multirow[c]{3}{*}{study-outcome} & 0.0 & $0.633 \pm 0.006$ & $0.653 \pm 0.005$ & $0.638 \pm 0.024$ & $\mathbf{0.657} \pm 0.004$ \\
     &  & 0.3 & $0.631 \pm 0.006$ & $0.651 \pm 0.005$ & $0.635 \pm 0.024$ & $\mathbf{0.656} \pm 0.004$ \\
     &  & 0.9 & $0.626 \pm 0.007$ & $0.646 \pm 0.004$ & $0.626 \pm 0.024$ & $\mathbf{0.651} \pm 0.005$ \\
    \bottomrule
\end{tabular}
}
    \caption{Performance of individual training regimes, trained on the complete training split, assessed
using temporally decayed AUROC with standard deviations (higher indicates better performance) over the
combined original validation and test splits.}
    \label{tab:decay-cls-std}
\end{table}

\newpage




\end{document}